\documentclass[conference,a4paper]{IEEEtran}
\IEEEoverridecommandlockouts

\usepackage[hidelinks]{hyperref}
\usepackage[cmex10]{amsmath}
\usepackage{amssymb,amsfonts}
\usepackage{dblfloatfix}
\usepackage[ruled,vlined]{algorithm2e}
\usepackage{graphicx}
\graphicspath{{Figures/PDF/}{Figures/PNG/}}
\usepackage{subcaption}
\usepackage{multirow}
\usepackage[table,xcdraw]{xcolor}
\usepackage{booktabs}
\usepackage{siunitx}
\usepackage[numbers,compress]{natbib}
\usepackage{texnames}
\usepackage{bm,bbm}
\usepackage{orcidlink}
\usepackage{xcolor}

\DeclareMathOperator*{\argmax}{arg\,max}

\begin{document}

\title{Fine-Tuning Adversarially-Robust Transformers for Single-Image Dehazing 
\thanks{This research was supported by \href{https://ai4risk.ro}{AI4RISK}, PN-IV-P6-6.3-SOL-2024-2-0251}
}

\author{	\IEEEauthorblockN{Vlad-Mihai Vasilescu \orcidlink{0000-0003-3743-4235}}
	\IEEEauthorblockA{\textit{\href{https://campus.pub.ro}{CAMPUS Research Institute}}\\
		\textit{Politehnica Bucharest}, Romania\\
		\href{mailto:vlad.vasilescu2111@upb.ro}{vlad.vasilescu2111@upb.ro}}
	\and
    
	\IEEEauthorblockN{Ana-Antonia Neacșu \orcidlink{0009-0003-4624-9442}}
	\IEEEauthorblockA{\textit{\href{https://campus.pub.ro}{CAMPUS Research Institute}}\\
		\textit{Politehnica Bucharest}, Romania\\
		\href{mailto:ana_antonia.neacsu@upb.ro}{ana\_antonia.neacsu@upb.ro}}
	\and
    
	\IEEEauthorblockN{Daniela Faur \orcidlink{0000-0001-5208-5753}}
	\IEEEauthorblockA{\textit{\href{https://campus.pub.ro}{CAMPUS Research Institute}}\\
		\textit{Politehnica Bucharest}, Romania\\
		\href{mailto:daniela.faur@upb.ro}{daniela.faur@upb.ro}}
}
\maketitle

\begin{abstract}
    Single-image dehazing is an important topic in remote sensing applications, enhancing the quality of acquired images and increasing object detection precision. However, the reliability of such structures has not been sufficiently analyzed, which poses them to the risk of imperceptible perturbations that can significantly hinder their performance. In this work, we show that state-of-the-art image-to-image dehazing transformers are susceptible to adversarial noise, with even 1 pixel change being able to decrease the PSNR by as much as 2.8 dB. Next, we propose two lightweight fine-tuning strategies aimed at increasing the robustness of pre-trained transformers. Our methods results in comparable clean performance, while significantly increasing the protection against adversarial data. We further present their applicability in two remote sensing scenarios, showcasing their robust behavior for out-of-distribution data.
    The source code for adversarial fine-tuning and attack algorithms can be found at \textcolor{blue}{\href{https://github.com/Vladimirescu/RobustDehazing}{Vladimirescu/RobustDehazing.git}}.
\end{abstract}


\section{Introduction}

Single-image dehazing is the task of estimating haze-free images from hazy observations. Given a haze-free image $x$, the degradation process is characterized as follows \cite{narasimhan2002vision}:
\begin{equation}
    y(i) = x(i) t(i) + A \big(1 - t(i)\big),\quad t(i) = e^{-\beta d(i)}
\end{equation}
where for each pixel $i$, $t(i)$ is the medium transmission map, $y$ is the hazy image, $A$ is the global atmospheric map, $\beta$ is the scattering coefficient, and $d(i)$ is the scene depth. Single-image dehazing is an ill-posed inverse problem that aims to estimate the clean image $x$ given only $y$. 
Early works focused on constraining the solution space and separately estimating $A$ and $t$ \cite{he2010single}. However, these multi-step approaches often yielded non-realistic haze-free predictions, and have been superseded by recent deep learning methods \cite{qiu2023mb, song2023vision, qin2020ffa, zheng2022dehaze}. These modern techniques leverage extensive training on synthetically constructed large-scale datasets to effectively predict haze-free images \cite{li2018benchmarking}.
In remote sensing, dehazing plays a crucial role in enhancing image quality \cite{zhu2021atmospheric}, enabling accurate object detection from UAVs \cite{feng2024HazyDet}, and improving object tracking capabilities \cite{sun2023adaptive}. 
Therefore, it is mandatory to create reliable systems capable of efficient dehazing in various scenarios.

Although neural networks are the most effective tools for dehazing, they are vulnerable to adversarial perturbations.
Recent adversarial attacks \cite{carlini2017towards, madry2018towards} have highlighted the failure of neural networks to withstand imperceptible perturbations, posing a high risk for remotely deployed systems. While protection methods for simple tasks such as classification continue to be developed \cite{cohen2019certified, neaccsu2024abba}, image-to-image (\textit{i2i}) networks lack the popularity of adversarial attacks and defenses. This poses a high risk for applications tasked with image restoration (e.g. super-resolution, dehazing), as it becomes easier to find efficient attacks, while the defense literature remains underdeveloped. Recently, the first image dehazing attack has been introduced \cite{gui2024fooling}, acting as a catalyst for the development of more advanced \textit{i2i} restoration attacks.

This paper investigates the robustness of single-image dehazing networks and aims to enhance the robustness of pre-trained architectures using two novel fine-tuning strategies, ultimately reducing the risk of adversarial manipulation in remote sensing autonomous systems. We present two \textit{i2i} adversarial attacks, review defense methods, outline our fine-tuning strategies, detail the setup and discuss the results.

\begin{figure}[t]
    \centering
    \includegraphics[width=0.67\linewidth]{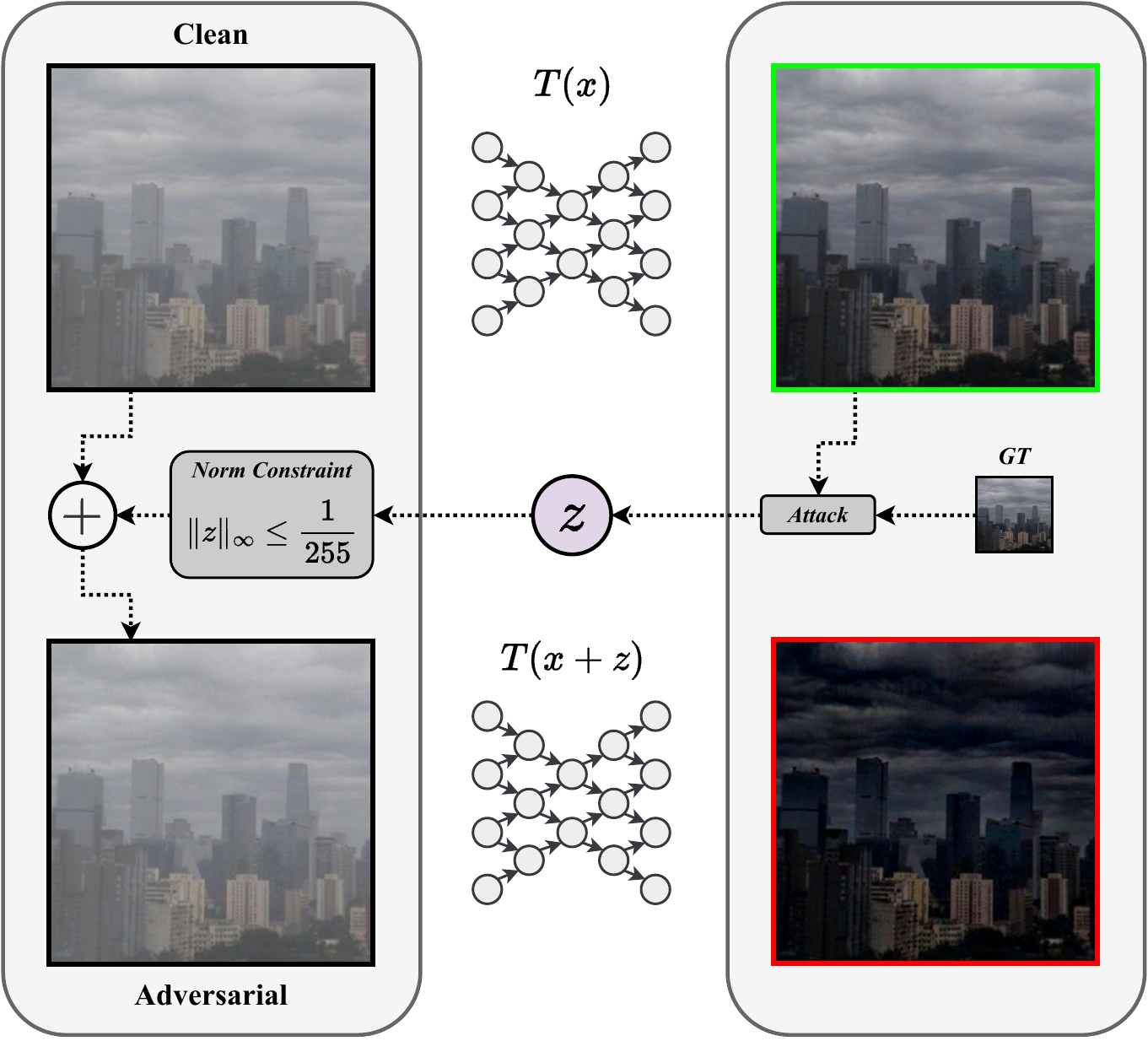}
    \caption{Small adversarial noise disrupts \textit{i2i} dehazing networks.}
    \label{fig:first}
    \vspace{-0.2cm}
\end{figure}

\section{Adversarial Attacks \& Robustness}

\subsection{Adversarial Attacks}

Given a neural network $T: \mathcal{D}_{in} \rightarrow \mathcal{D}_{out}$ and input-output pair $(x, y) \in \mathcal{D}_{in} \times \mathcal{D}_{out}$, we are interested in finding perturbation $z$ which maximally fools $T$ under budget $\epsilon$: 
\begin{equation}
    z = \argmax_{\delta, \, \Vert \delta \Vert_p \leq \epsilon} \mathcal{L}\Big(T(x), T(x + \delta), y\Big)
\end{equation}
where $\mathcal{L}$ is a search objective and $p \geq 0$. Usually, the perturbation $z$ is found by maximizing some measure of dissimilarity between $T(x)$ and $T(x+\delta)$. In classification tasks, $T(x)$ corresponds to the logit vector and an attack is considered successful when the logit of highest value is replaced by another. 

For image-predicting networks, however, there is no general formula for deeming an attack as successful. In this paper, our focus is on networks directly predicting dehazed images, which naturally raises the objective of maximizing the dissimilarity between $T(x + z)$ and $y$:
\begin{equation}\label{eq:attack}
    z = \argmax_{\delta, \Vert \delta \Vert_p \leq \epsilon} \Vert T(x + \delta) - y \Vert_q
\end{equation}
with $p, q \geq 0$. We choose $q=1$ in all our experiments. 
This can be iteratively solved through standard first-order gradient methods. In this study, we focus on the case $p=\infty$ (which we term $\ell_\infty$-attack), since $\Vert \delta \Vert_\infty$ can be easily interpreted. This corresponds to the widely-known PGD-attack \cite{madry2018towards} that served as a base for the majority of classification attacks to come. For extensive examples regarding attacks on various dehazing strategies check out \cite{gui2024fooling}.

Another prominent attack strategy is the OnePixel attack \cite{su2019one}, which modifies only one input pixel to maximally perturb the system. We modified this attack to work on \textit{i2i} dehazing networks (which we termed $\ell_0$-attack), changing the original search objective with the one from Eq.\ref{eq:attack}. Although not as powerful as the $\ell_\infty$-attack, $\ell_0$-attack will be shown to have a strong effect even when changing only $1 / 256^2$ input pixels.

\subsection{Robust Training} 

To overcome the powerful effect of adversarial noise, several techniques were developed to promote training of more robust structures \cite{carlini2017towards, serrurier2021achieving}. Three main categories of robust training techniques were showcased in previous works: adversarial training, regularization and certified methods.

Adversarial Training (\texttt{AT}) \cite{madry2018towards} adapts the network to adversarial data by incorporating adversarial noise into training samples. Consider a dataset of size $N$:
\begin{equation}
    \mathcal{D}_{\text{train}} = \{x_i, y_i\}_{i=1}^N \rightarrow \mathcal{D}'_{\text{train}} = \{\Pi_{\mathcal{D}_{in}} (x_i + z_i), y_i\}_{i=1}^N,
\end{equation}
where $\mathcal{D}'_{\text{train}}$ can be re-generated either every epoch or individually for each batch at iteration level. Although straightforward,
this remains the state-of-the-art robust training method in terms of accuracy-robustness trade-off. 

Regularization techniques differ by adding additional loss terms to guide the optimization process towards more robust models. \texttt{TRADES} \cite{zhang2019theoretically} adds a term to increase the similarity between predictions over clean and adversarial data:
\begin{equation}
    \mathcal{L} = \mathcal{L}_{\text{base}} + \lambda \mathcal{L}_{reg}\big(T(x_i), T(x_i + z_i)\big)
\end{equation}
where $\lambda > 0$ is a regularization hyperparameter and $\mathcal{L}_{\text{reg}}$ encodes some distance between the two predictions.  

The last category of certified methods encompasses techniques which specifically impose robustness under a certain level of perturbation \cite{cohen2019certified}. They work by constraining the network structure s.t. different certifications over predictions can be formulated \cite{neaccsu2024abba, pauli2021training}. However, these strategies come with a high computational overhead, which limits the size and complexity of the network. For this reason, we exclude them from this study and focus only on the first two categories, leaving their usage in dehazing as a future endeavor.

\section{Methodology}

\begin{table}[t]
\centering
\resizebox{0.8\linewidth}{!}{\begin{tabular}{cccc}
\toprule
\textbf{Architecture}                       & \textbf{Method} & \textbf{Params.} (M) & \textbf{Tuned} (\%) \\ \midrule
\multirow{3}{*}{\texttt{DehazeFormer-T}}    
                                   & \texttt{LL}                 & $\approx 0.68$       & 1.54                   \\
                                   & \texttt{SB}                 & $\approx 0.68$       & 0.12                   \\
                                   & \texttt{LINEAD}             & 1.2              & 41.5                   \\ \midrule
\multirow{3}{*}{\texttt{MB-TaylorFormer-B}} 
                                   & \texttt{LL}                 & $\approx 2.7$        & 0.05                   \\
                                   & \texttt{SB}                 & $\approx 2.7$        & 0.006                  \\
                                   & \texttt{LINEAD}             & 3.2              & 16.5                   \\ \midrule
\multirow{3}{*}{\texttt{DehazeFormer-B}}   
                                   & \texttt{LL}                 & $\approx 2.5$        & 1.55                   \\
                                   & \texttt{SB}                 & $\approx 2.5$        & 0.03                   \\
                                   & \texttt{LINEAD}             & $4.5$        & 43.63 \\ \bottomrule                
\end{tabular}}
\caption{Stats for each network and method.}
\label{tab:stats}
    \vspace{-0.7em}
\end{table}

\begin{figure}
    \centering
    \includegraphics[width=0.8\linewidth]{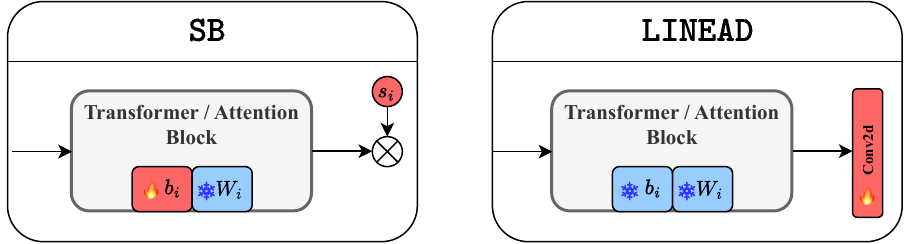}
    \caption{Our two proposed fine-tuning methods.}
    \label{fig:blocks}
    \vspace{-0.3cm}
\end{figure}

\begin{table*}[t]
\resizebox{\textwidth}{!}{\begin{tabular}{ccccccccccccccc}
\toprule \toprule
\multicolumn{1}{c}{} & \multicolumn{2}{c}{} & \multicolumn{2}{c}{} & \multicolumn{2}{c}{}& \multicolumn{8}{c}{\textbf{Adversarial Attacks}} \\ \cmidrule{8-15} 

\multicolumn{1}{c}{} & \multicolumn{2}{c}{} & \multicolumn{2}{c}{} & \multicolumn{2}{c}{}& \multicolumn{4}{c}{$\ell_{\infty}$-attack}& \multicolumn{4}{c}{$\ell_{0}$-attack} \\ \cmidrule(lr){8-11} \cmidrule(lr){12-15} 

\multicolumn{1}{c}{} & \multicolumn{2}{c}{\multirow{-4}{*}{\textbf{Fine-Tuning Setup}}} & \multicolumn{2}{c}{\multirow{-4}{*}{\textbf{Clean Data}}} & \multicolumn{2}{c}{\multirow{-4}{*}{\textbf{\begin{tabular}[c]{@{}c@{}} Gaussian Noise \\ $z \sim \mathcal{N}(0, 0.01)$ \\$\Vert z\Vert_{\infty} \approx \frac{12}{255}$ \end{tabular}}}} & \multicolumn{2}{c}{$\Vert z\Vert_{\infty} \leq \frac{1}{255}$} & \multicolumn{2}{c}{$\Vert z\Vert_{\infty} \leq \frac{4}{255}$}& \multicolumn{2}{c}{$\Vert z\Vert_{0} = 1$ pixel}& \multicolumn{2}{c}{$\Vert z\Vert_{0} = 8$ pixels} \\ \cmidrule{2-3} \cmidrule(lr){4-5} \cmidrule(lr){6-7} \cmidrule(lr){8-9} \cmidrule(lr){10-11} \cmidrule(lr){12-13} \cmidrule(lr){14-15}

\multicolumn{1}{c}{\multirow{-6}{*}{\textbf{Architecture}}} & \textbf{Method} & \textbf{Defense} &  \textbf{PSNR} & \textbf{SSIM} & \textbf{PSNR} & \textbf{SSIM} & \textbf{PSNR} & \textbf{SSIM} & \textbf{PSNR} & \textbf{SSIM} & \multicolumn{1}{c}{\textbf{PSNR}} & \multicolumn{1}{c}{\textbf{SSIM}} & \multicolumn{1}{c}{\textbf{PSNR}} & \multicolumn{1}{c}{\textbf{SSIM}}\\ \midrule 
 
& \multicolumn{2}{c}{Base} & 33.71 & 0.982 & 33.01 & 0.956 & 16.26 & 0.792 & 13.35 & 0.634 & \multicolumn{1}{c}{32.65} & \multicolumn{1}{c}{0.967} & \multicolumn{1}{c}{29.23} & \multicolumn{1}{c}{0.959} \\ \cmidrule{2-15}

&\texttt{LL}& \texttt{AT} & 27.72 & 0.934 & 27.34 & 0.921 & 19.46 & 0.851 & 17.89 & 0.713 & 27.4 & 0.933 & \multicolumn{1}{c}{26.41} & \multicolumn{1}{c}{0.926}\\

&\texttt{SB}& \texttt{AT} & 29.08 & 0.955 & 28.90& 0.943 & 25.44 & 0.932 & 19.11 & 0.803 & 28.96 & 0.954 & \multicolumn{1}{c}{28.34} & \multicolumn{1}{c}{0.950} \\

& \texttt{LINEAD} & \texttt{AT} & 31.48 & 0.963 & 31.21 & 0.954 & \textbf{26.54} & \textbf{0.940} & \underline{19.44} & \underline{0.815} & 31.21 & 0.962 & \underline{29.82} & \multicolumn{1}{c}{0.957}\\

& \texttt{LINEAD} & \texttt{TRADES}$_{\lambda = 1.0}$ & 30.92 & \underline{0.969} & 30.68 & 0.953 & \underline{26.47} & \underline{0.939} & \textbf{19.80} & \textbf{0.822} & 30.71 & 0.961 & \multicolumn{1}{c}{29.52} & \multicolumn{1}{c}{0.956}\\

& \texttt{LINEAD} & \texttt{TRADES}$_{\lambda = 0.5}$& \underline{32.19} & 0.965 & \underline{31.74} & \underline{0.955} & 23.48 & 0.916 & 16.69 & 0.741 & \underline{31.78} & \underline{0.964} & \textbf{30.92} & \textbf{0.961} \\

\multirow{-7}{*}{\footnotesize \texttt{DehazeFormer-T}} & \texttt{LINEAD} & \texttt{TRADES}$_{\lambda = 0.1}$ & \textbf{33.65} & \textbf{0.969} & \textbf{32.88} & \textbf{0.958} & 17.06 & 0.808 & 13.73 & 0.619 & \textbf{32.58} & \textbf{0.967} & \multicolumn{1}{c}{29.37} & \underline{0.959} \\ \midrule

& \multicolumn{2}{c}{Base} & 33.78 & 0.969 & 33.04 & 0.954 & 15.81 & 0.751 & 14.8 & 0.585 & 31.35 & 0.964 & 28.33 & 0.955 \\ \cmidrule{2-15}

 &\texttt{LL}& \texttt{AT} & \underline{31.60} & 0.953 & \underline{31.23} & 0.938 & 16.41 & 0.782 & 15.53 & 0.583 & 30.13 & 0.947 & 27.93 & 0.938 \\
 
 &\texttt{SB}& \texttt{AT} & \textbf{33.21} & \textbf{0.967} & \textbf{32.81} & \textbf{0.952} & 16.44 & 0.800& 15.07 & 0.627 & \textbf{31.45} & \textbf{0.962} & 28.34 & 0.953 \\
 
& \texttt{LINEAD} & \texttt{AT} & 30.47 & \underline{0.960} & 30.17 & \underline{0.946} & \underline{26.08} & \underline{0.937} & \underline{19.48} & \textbf{0.802} & \underline{30.17} & \underline{0.959} & \textbf{28.95} & \textbf{0.955} \\

\multirow{-5}{*}{\footnotesize \texttt{MB-TaylorFormer}}& \texttt{LINEAD} & \texttt{TRADES}$_{\lambda = 0.5}$ & 30.39 & \underline{0.960} & 30.08 & \underline{0.946} & \textbf{26.26} & \textbf{0.937} & \textbf{19.53} & \underline{0.797} & 30.14 & \underline{0.959} & \underline{28.89} & \underline{0.955} \\ \midrule

& \multicolumn{2}{c}{Base} & 34.86 & 0.971 & 33.42 & 0.959 & 17.56 & 0.833 & 14.17 & 0.629 & 32.02 & 0.968 & 29.16 & 0.962 \\ \cmidrule{2-15}

&\texttt{LL}& \texttt{AT} & 30.27 & 0.949 & 29.54 & 0.935 & 18.41 & 0.844 & 16.84 & 0.672 & 29.13 & 0.946 & 27.77 & 0.940 \\

&\texttt{SB}& \texttt{AT} & 29.64 & 0.958 & 29.39 & 0.948 & \underline{25.36} & \underline{0.934} & \underline{18.65} & \underline{0.793} & 29.39 & 0.957 & 28.66 & 0.955 \\

& \texttt{LINEAD} & \texttt{AT} & \underline{31.13} & \underline{0.963}  & \underline{30.85} & \underline{0.955} & \textbf{26.46} & \textbf{0.942} & \textbf{19.49} & \textbf{0.829} & \underline{30.67} & \underline{0.962} & \underline{28.91} & \underline{0.957} \\

\multirow{-5}{*}{\footnotesize \texttt{DehazeFormer-B}}& \texttt{LINEAD} & \texttt{TRADES}$_{\lambda = 0.5}$ & \textbf{32.86} & \textbf{0.967} & \textbf{32.41} & \textbf{0.958} & 24.84 & 0.927 & 17.82 & 0.773 & \textbf{32.22} & \textbf{0.966} & \textbf{29.76} & \textbf{0.961} \\ \bottomrule \bottomrule
\end{tabular}}
\caption{Robustness against different adversarial noise $z$ for base and fine-tuned models on the RESIDE-Outdoor test set.}
\label{tab:results}
    \vspace{-0.7em}
\end{table*}

\begin{figure*}[t]
    \centering
    \includegraphics[width=\textwidth]{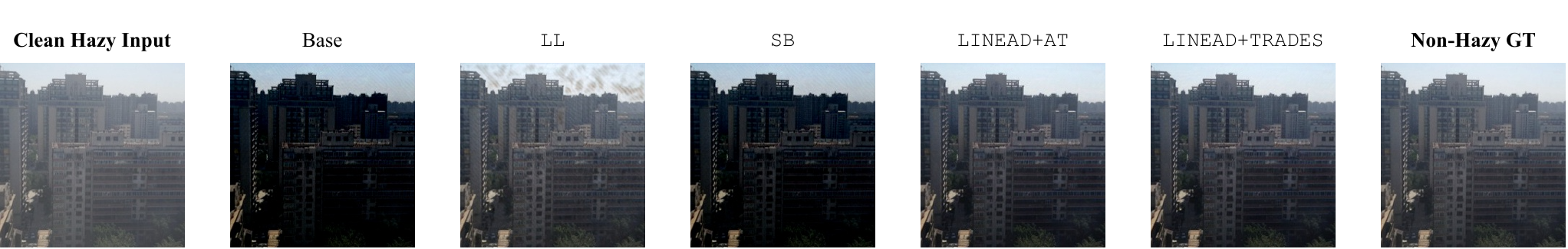}
    \caption{Dehazing results for a clean input sample attacked with $\Vert z \Vert_{\infty} \leq \frac{1}{255}$, using different \texttt{MB-TaylorFormer-B} models.}
    \label{fig:sample}
        \vspace{-0.7em}
\end{figure*}

\begin{figure*}[!t]
    \centering
    \includegraphics[width=\textwidth]{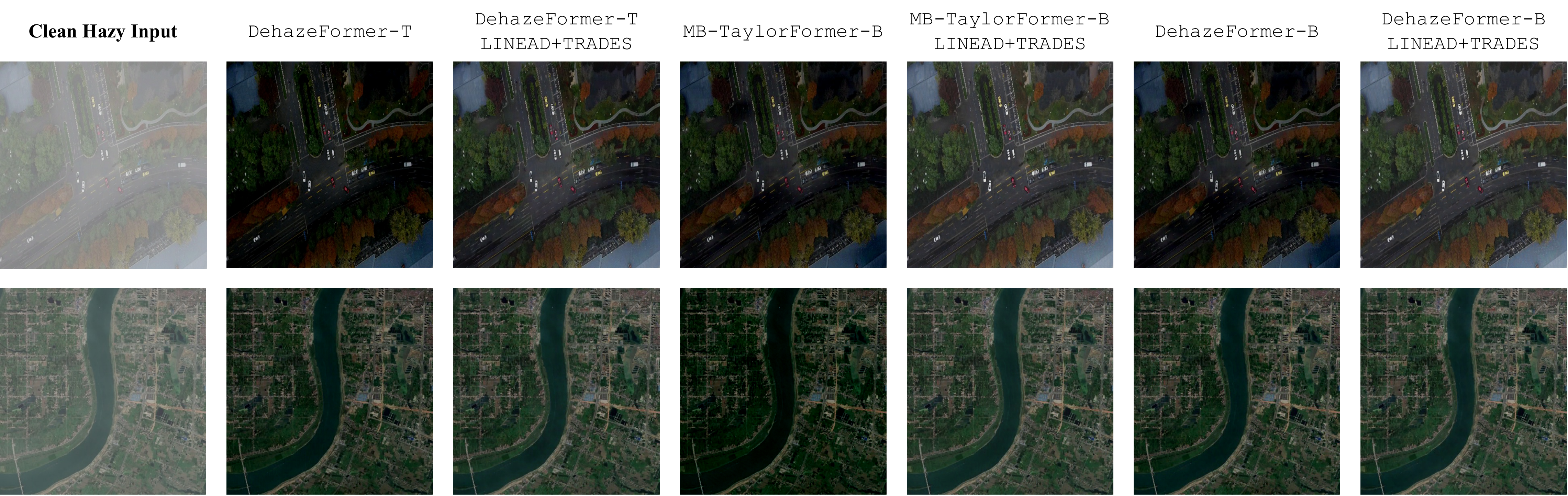}
    \caption{Dehazing results for two samples from HazyDet UAV (first row) and RICE-I (second row) attacked with $\Vert z \Vert_{\infty} \leq \frac{1}{255}$.}
    \label{fig:drone}
        \vspace{-0.7em}
\end{figure*}

\begin{table}[!t]
\centering
\resizebox{0.85\linewidth}{!}{\begin{tabular}{cccccc}
\toprule \toprule
\multicolumn{2}{c}{}  & \multicolumn{2}{c}{\textbf{Clean Data}} & \multicolumn{2}{c}{\textbf{$\Vert z\Vert_{\infty} = \frac{1}{255}$}}\\ \cmidrule(lr){3-4} \cmidrule(lr){5-6}

\multicolumn{2}{c}{\multirow{-3}{*}{\textbf{Model}}}& \textbf{PSNR}  & \textbf{SSIM} & \textbf{PSNR}  & \textbf{SSIM} \\ \midrule

\multicolumn{6}{c}{\textbf{HazyDet} (\texttt{PSNR = 14.90})} \\ \midrule

& Base & 19.99 & 0.846 & \textcolor{red}{12.82} & 0.676\\

\multirow{-2}{*}{\texttt{DehazeFormer-T}} & \texttt{LINEAD + TRADES}  & 19.70 & 0.838& 16.22 & 0.761\\ \midrule

& Base & 19.97 & 0.831& \textcolor{red}{12.86} & 0.647  \\

\multirow{-2}{*}{\texttt{MB-TaylorFormer}} & \texttt{LINEAD + TRADES}  & 18.94 & 0.828 & 16.96 & 0.784\\ \midrule

& Base & 20.27 & 0.850 & \textcolor{red}{12.91} & 0.658\\

\multirow{-2}{*}{\texttt{DehazeFormer-B}} & \texttt{LINEAD + TRADES}  & 19.91 & 0.842 & 16.77 & 0.733  \\ \midrule 

\multicolumn{6}{c}{\textbf{RICE-I} (\texttt{PSNR = 17.04})} \\ \midrule


& Base & 21.64 & 0.875 & \textcolor{red}{13.02} & 0.420\\

\multirow{-2}{*}{\texttt{DehazeFormer-T}} & \texttt{LINEAD + TRADES}  & 22.24 & 0.890 & 17.08 & 0.789 \\ \midrule

& Base & 23.36 & 0.884 & \textcolor{red}{13.71} & 0.436  \\

\multirow{-2}{*}{\texttt{MB-TaylorFormer}} & \texttt{LINEAD + TRADES}  & 20.07 & 0.876 & 18.22 & 0.830 \\ \midrule

& Base & 21.86 & 0.881 & \textcolor{red}{14.73} & 0.636 \\

\multirow{-2}{*}{\texttt{DehazeFormer-B}} & \texttt{LINEAD + TRADES}  & 22.47 & 0.898 & \textcolor{red}{16.90} & 0.768 \\ \bottomrule \bottomrule
\end{tabular}}

\caption{Results on HazyDet UAV and RICE-I datasets. \textcolor{red}{Red} indicates a drop below the original \texttt{PSNR} value.}
\label{tab:hazydet}
    \vspace{-0.8em}
\end{table}

\subsection{Fine-tuning for robustness}

As initially stated, we focus on fine-tuning  pre-trained models to enhance their robustness against \textit{i2i} adversarial attacks \cite{li2021improved}. This approach seeks to eliminate the necessity to re-train models from scratch to ensure robustness, which may introduce various bottlenecks.

To this end, we propose the following fine-tuning strategies:

\begin{itemize}
    \item \textit{Last Layer} (\textbf{\texttt{LL}}), which only trains the final layer of the network, keeping all the others frozen. This method was proven efficient in multiple classification tasks when distribution changes occur in the input space.
   
    \item \textit{Scale-and-Bias} (\textbf{\texttt{SB}}). Inspired by \textit{DiffFit} \cite{xie2023difffit} used in Diffusion models, we add a trainable scaling factor after each Transformer / Attention block in the architecture, initialized with $1.0$. Additionally, we let bias vectors be trainable, keeping all the other parameters frozen (Fig.\ref{fig:blocks}). As the bias parameters account for a small percent of the total, \texttt{SB} is very lightweight and easily adapts to data distribution changes, due to the new scaling factors which act as pseudo-normalizers.

    \item \textit{LINEar ADaptation} (\textbf{\texttt{LINEAD}}), where we add new trainable linear layers after each Transformer / Attention block, maintaining the output dimensionality (Fig.\ref{fig:blocks}). In order to avoid chaotic initial behaviors that would hinder the frozen model performance, we initialize these layers with identity operators, starting from a similar performance in the first iterations and gradually adapting to adversarial noise through these new layers. 
\end{itemize}


We use these fine-tuning methods together with robust training techniques to create lightweight tuning strategies, yielding robust structures against adversarial attacks.


\subsection{Measuring Robustness}
To assess the robustness of base / fine-tuned models against adversarial noise, we employ the previously proposed $\ell_\infty$ and $\ell_0$ attacks. For the $\ell_\infty$ attack, we evaluate with $\epsilon = \frac{1}{255}$ and $\epsilon = \frac{4}{255}$, indicating that each pixel can vary by at most these amounts. These low values are chosen to ensure that the adversarially perturbed image remains perceptually similar to the clean image, as illustrated in Fig.\ref{fig:first}.
For the $\ell_0$-attack, we evaluate the performance drop when changing either only 1 pixel or 8 pixels in the input image. As comparison, we evaluate the influence of plain Gaussian noise added over input data, using $\sigma = 0.01$ ($ \Vert z\Vert_\infty \approx \frac{12}{255}$) to emphasize the effectiveness of our attacks.

\section{Experiments \& Results}

\subsection{Setup}

We choose three high-performance Transformer models for dehazing as our base architectures, namely \texttt{MB-TaylorFormer-B} \cite{qiu2023mb}, \texttt{DehazeFormer-T}  and \texttt{DehazeFormer-B} \cite{song2023vision}. The latter two are different configurations of the same base model, and are chosen to illustrate that robustness depends on the network size scale.

For fine-tuning and evaluation, we select the widely used RESIDE-Outdoor \cite{li2018benchmarking} train and test sets. We do not use the Indoor set, arguing that haze distributions are significantly different and the models cannot generalize further to remote sensing applications. Fine-tuned models were additionally tested on two remote sensing dehazing datasets:
HazyDet \cite{feng2024HazyDet}, a recently introduced dataset containing images acquired from UAVs for dehazing and object detection, and RICE-I \cite{lin2019remote}, with images acquired from Google Earth. We use 50 images from both of these datasets for testing, at their original scale. Our goal is to show that models fine-tuned on a large-scale outdoor dataset can be robust to out-of-distribution remote-sensing data. Therefore, we employ as starting checkpoints the models trained solely on the RESIDE-Outdoor subset.

For both \texttt{AT} and \texttt{TRADES} we use $\ell_\infty$-attack with $\Vert z\Vert_\infty = \frac{1}{255}$ to craft adversarial examples during training. We generate these samples at the iteration level, \textit{i.e.} during each batch iteration. 
We did not assess the impact of training under $\ell_0$-attacks because of the substantial computational cost associated with their execution.
For \texttt{TRADES}, we tested multiple values for $\lambda$, finding that $0.5$ worked best. 
For \texttt{LINEAD}, the newly added layers have $3\times 3$ \texttt{Conv2d} filters initialized with identity, and bias vectors initialized with $0$.

Models are fine-tuned using $256\times 256$ patches. As performance measures, we use the classic PSNR and SSIM. For training we employed the Adam optimizer with a learning rate of $10^{-5}$, along with the $\ell_1$ loss. To further reduce the computational cost and training time, we use a random sampler to draw at each epoch $5000$ input patches for fine-tuning. All models undergo fine-tuning for $15$ epochs, using different batch sizes for each method depending on GPU memory availability. Parameter statistics for each configuration are reported in Table \ref{tab:stats}. Given these settings, we use a single NVIDIA A6000 GPU for all our experiments.

\subsection{Results}

Numerical results obtained on the RESIDE-Outdoor test set are reported in Table \ref{tab:results}. 
The baseline \texttt{LL} method exhibited inferior performance compared to the two proposed methods, which achieved good performance on clean (non-adversarial) data. This shows that training only one layer is not sufficient to adequately model the structural characteristics of the adversarial noise.
\texttt{SB} and \texttt{LINEAD} allow the newly formed architecture to easily adapt to adversarial data, thus, increasing its robustness. Gaussian Noise has a reduced effect over performance, even for a higher $\ell_\infty$ magnitude, showing the effectiveness of our adversarial attacks.

In terms of $\ell_\infty$-attack results, \texttt{LINEAD} leads to the most robust models, in all 3 cases. However, it comes with the drawback of a heavy computational overhead, due to the large number of parameters being fine-tuned (check out the percentages from Table \ref{tab:stats}). \texttt{SB} on the other hand, yields slightly lower \texttt{PSNR} at attacks, but it comes with the benefit of a fast fine-tuning, and could be easily applied to more complex structures. For \texttt{DehazeFormer-T} we illustrate the influence of $\lambda$ during \texttt{TRADES}, acting as a trade-off parameter between clean and robust performance. This shows that we can obtain comparable performance on clean data, while increasing the robustness profile. A visual example is shown in Fig.\ref{fig:sample}.

In terms of $\ell_0$-attack, all base models seem to be susceptible to even 1 pixel change, with \texttt{PSNR} dropping by at least \texttt{1 dB} and at most \texttt{2.84 dB} (for \texttt{DehazeFormer-B}). This indicates that large variants of the same architecture can be more easily fooled, due to their extensive representation landscape which leaves room for more complex noise. 

For the two remote-sensing datasets, results are shown in Table \ref{tab:hazydet}. It is clear that even for a minor decrease in performance on clean data, models robustly fine-tuned on RESIDE offer increased protection against adversarial attacks, with the base models getting fooled up to the point of a \texttt{PSNR} lower than the original one. This goes to show that models trained and fine-tuned solely on outdoor synthetic data can still generalize well to adversarial samples for out-of-distribution data. 
Predictions over two examples from HazyDet and RICE-I, attacked with $\Vert z\Vert_\infty \leq \frac{1}{255}$, are shown in Fig.\ref{fig:drone}. This reflects the behavior on the RESIDE-Outdoor distribution, emphasizing the generalizable nature of our fine-tuned models.
 
\section{Conclusions}

In this paper, we tackled the possibility of increasing robustness for single-image dehazing transformers through two proposed lightweight fine-tuning methods, namely \texttt{SB} and \texttt{LINEAD}. Using our $\ell_\infty$ and $\ell_0$ attacks, we showed that robustly fine-tuned models exert increased protection against adversarial data, while maintaining good clean performance. These models were also shown to perform well on remote sensing out-of-distribution data, both on clean and adversarial images. To the best of our knowledge, this is one of the first works to analyze the robustness of neural networks specialized in dehazing. We plan on expanding these results, analyzing more closely our proposed fine-tuning methods and going further with their usage in different applications.

\newpage
\clearpage

\small


\begin{thebibliography}{22}
\providecommand{\natexlab}[1]{#1}
\providecommand{\url}[1]{#1}
\csname url@samestyle\endcsname
\providecommand{\newblock}{\relax}
\providecommand{\bibinfo}[2]{#2}
\providecommand{\BIBentrySTDinterwordspacing}{\spaceskip=0pt\relax}
\providecommand{\BIBentryALTinterwordstretchfactor}{4}
\providecommand{\BIBentryALTinterwordspacing}{\spaceskip=\fontdimen2\font plus
\BIBentryALTinterwordstretchfactor\fontdimen3\font minus \fontdimen4\font\relax}
\providecommand{\BIBforeignlanguage}[2]{{%
\expandafter\ifx\csname l@#1\endcsname\relax
\typeout{** WARNING: IEEEtranN.bst: No hyphenation pattern has been}%
\typeout{** loaded for the language `#1'. Using the pattern for}%
\typeout{** the default language instead.}%
\else
\language=\csname l@#1\endcsname
\fi
#2}}
\providecommand{\BIBdecl}{\relax}
\BIBdecl

\bibitem[Narasimhan and Nayar(2002)]{narasimhan2002vision}
S.~G. Narasimhan and S.~K. Nayar, ``Vision and the atmosphere,'' \emph{International Journal of Computer Vision}, vol.~48, pp. 233--254, 2002.

\bibitem[He et~al.(2010)He, Sun, and Tang]{he2010single}
K.~He, J.~Sun, and X.~Tang, ``Single image haze removal using dark channel prior,'' \emph{IEEE Transactions on Pattern Analysis and Machine Intelligence}, vol.~33, no.~12, pp. 2341--2353, 2010.

\bibitem[Qiu et~al.(2023)Qiu, Zhang, Wang, Luo, Li, and Jin]{qiu2023mb}
Y.~Qiu, K.~Zhang, C.~Wang, W.~Luo, H.~Li, and Z.~Jin, ``Mb-taylorformer: {M}ulti-branch efficient transformer expanded by taylor formula for image dehazing,'' in \emph{Proc. International Conference on Computer Vision}, 2023, pp. 12\,802--12\,813.

\bibitem[Song et~al.(2023)Song, He, Qian, and Du]{song2023vision}
Y.~Song, Z.~He, H.~Qian, and X.~Du, ``Vision transformers for single image dehazing,'' \emph{IEEE Transactions on Image Processing}, vol.~32, pp. 1927--1941, 2023.

\bibitem[Qin et~al.(2020)Qin, Wang, Bai, Xie, and Jia]{qin2020ffa}
X.~Qin, Z.~Wang, Y.~Bai, X.~Xie, and H.~Jia, ``{FFA-Net}: {F}eature fusion attention network for single image dehazing,'' in \emph{Proc. AAAI Conference on Artificial Intelligence}, vol.~34, no.~07, 2020, pp. 11\,908--11\,915.

\bibitem[Zheng et~al.(2022)Zheng, Su, Zhang, Tao, and Wang]{zheng2022dehaze}
Y.~Zheng, J.~Su, S.~Zhang, M.~Tao, and L.~Wang, ``{Dehaze-AGGAN}: {U}npaired remote sensing image dehazing using enhanced attention-guide generative adversarial networks,'' \emph{IEEE Transactions on Geoscience and Remote Sensing}, vol.~60, pp. 1--13, 2022.

\bibitem[Li et~al.(2018)Li, Ren, Fu, Tao, Feng, Zeng, and Wang]{li2018benchmarking}
B.~Li, W.~Ren, D.~Fu, D.~Tao, D.~Feng, W.~Zeng, and Z.~Wang, ``Benchmarking single-image dehazing and beyond,'' \emph{IEEE Transactions on Image Processing}, vol.~28, no.~1, pp. 492--505, 2018.

\bibitem[Zhu et~al.(2021)Zhu, Luo, Wei, Li, Qi, Mazur, Li, and Li]{zhu2021atmospheric}
Z.~Zhu, Y.~Luo, H.~Wei, Y.~Li, G.~Qi, N.~Mazur, Y.~Li, and P.~Li, ``Atmospheric light estimation based remote sensing image dehazing,'' \emph{Remote Sensing}, vol.~13, no.~13, p. 2432, 2021.

\bibitem[Feng et~al.(2024)Feng, Chen, Kou, Gao, Wang, Li, Shu, Dai, Fu, and Yang]{feng2024HazyDet}
C.~Feng, Z.~Chen, R.~Kou, G.~Gao, C.~Wang, X.~Li, X.~Shu, Y.~Dai, Q.~Fu, and J.~Yang, ``Hazy{D}et: {O}pen-source benchmark for drone-view object detection with depth-cues in hazy scenes,'' \emph{preprint arXiv:2409.19833}, 2024.

\bibitem[Sun et~al.(2023)Sun, Chang, Zhang, Fan, and He]{sun2023adaptive}
L.~Sun, J.~Chang, J.~Zhang, B.~Fan, and Z.~He, ``Adaptive image dehazing and object tracking in {UAV} videos based on the template updating siamese network,'' \emph{IEEE Sensors Journal}, vol.~23, no.~11, pp. 12\,320--12\,333, 2023.

\bibitem[Carlini and Wagner(2017)]{carlini2017towards}
N.~Carlini and D.~Wagner, ``Towards evaluating the robustness of neural networks,'' in \emph{IEEE Symposium on Security and Privacy}, 2017, pp. 39--57.

\bibitem[Madry et~al.(2018)Madry, Makelov, Schmidt, Tsipras, and Vladu]{madry2018towards}
A.~Madry, A.~Makelov, L.~Schmidt, D.~Tsipras, and A.~Vladu, ``Towards deep learning models resistant to adversarial attacks,'' in \emph{Proc. International Conference on Learning Representations}, 2018.

\bibitem[Cohen et~al.(2019)Cohen, Rosenfeld, and Kolter]{cohen2019certified}
J.~Cohen, E.~Rosenfeld, and Z.~Kolter, ``Certified adversarial robustness via randomized smoothing,'' in \emph{Proc. International Conference on Machine Learning}, 2019, pp. 1310--1320.

\bibitem[Neac{\c{s}}u et~al.(2024)Neac{\c{s}}u, Pesquet, Vasilescu, and Burileanu]{neaccsu2024abba}
A.~Neac{\c{s}}u, J.-C. Pesquet, V.~Vasilescu, and C.~Burileanu, ``{ABBA} neural networks: {C}oping with positivity, expressivity, and robustness,'' \emph{SIAM Journal on Mathematics of Data Science}, vol.~6, no.~3, pp. 649--678, 2024.

\bibitem[Gui et~al.(2024)Gui, Cong, Peng, Tang, and Kwok]{gui2024fooling}
J.~Gui, X.~Cong, C.~Peng, Y.~Y. Tang, and J.~T.-Y. Kwok, ``Fooling the image dehazing models by first order gradient,'' \emph{IEEE Transactions on Circuits and Systems for Video Technology}, vol.~34, no.~7, p. 6265–6278, 2024.

\bibitem[Su et~al.(2019)Su, Vargas, and Sakurai]{su2019one}
J.~Su, D.~V. Vargas, and K.~Sakurai, ``One pixel attack for fooling deep neural networks,'' \emph{IEEE Transactions on Evolutionary Computation}, vol.~23, no.~5, pp. 828--841, 2019.

\bibitem[Serrurier et~al.(2021)Serrurier, Mamalet, Gonz{\'a}lez-Sanz, Boissin, Loubes, and Del~Barrio]{serrurier2021achieving}
M.~Serrurier, F.~Mamalet, A.~Gonz{\'a}lez-Sanz, T.~Boissin, J.-M. Loubes, and E.~Del~Barrio, ``Achieving robustness in classification using optimal transport with hinge regularization,'' in \emph{Proc. IEEE/CVF Conference on Computer Vision and Pattern Recognition}, 2021, pp. 505--514.

\bibitem[Zhang et~al.(2019)Zhang, Yu, Jiao, Xing, El~Ghaoui, and Jordan]{zhang2019theoretically}
H.~Zhang, Y.~Yu, J.~Jiao, E.~Xing, L.~El~Ghaoui, and M.~Jordan, ``Theoretically principled trade-off between robustness and accuracy,'' in \emph{Proc. International Conference on Machine Learning}, 2019, pp. 7472--7482.

\bibitem[Pauli et~al.(2021)Pauli, Koch, Berberich, Kohler, and Allg{\"o}wer]{pauli2021training}
P.~Pauli, A.~Koch, J.~Berberich, P.~Kohler, and F.~Allg{\"o}wer, ``Training robust neural networks using {L}ipschitz bounds,'' \emph{IEEE Control Systems Letters}, vol.~6, pp. 121--126, 2021.

\bibitem[Li and Zhang(2021)]{li2021improved}
D.~Li and H.~Zhang, ``Improved regularization and robustness for fine-tuning in neural networks,'' vol.~34, 2021, pp. 27\,249--27\,262.

\bibitem[Xie et~al.(2023)Xie, Yao, Shi, Liu, Zhou, Liu, Li, and Li]{xie2023difffit}
E.~Xie, L.~Yao, H.~Shi, Z.~Liu, D.~Zhou, Z.~Liu, J.~Li, and Z.~Li, ``Difffit: {U}nlocking transferability of large diffusion models via simple parameter-efficient fine-tuning,'' in \emph{Proc. International Conference on Computer Vision}, 2023, pp. 4230--4239.

\bibitem[Lin et~al.(2019)Lin, Xu, Wang, Wang, Sun, and Fu]{lin2019remote}
D.~Lin, G.~Xu, X.~Wang, Y.~Wang, X.~Sun, and K.~Fu, ``A remote sensing image dataset for cloud removal,'' \emph{preprint arXiv:1901.00600}, 2019.

\end{thebibliography}
\end{document}